\documentclass[fleqn,10pt]{wlscirep}
\usepackage[utf8]{inputenc}
\usepackage[T1]{fontenc}
\usepackage{lineno}
\usepackage{cite}
\usepackage[separate-uncertainty = true]{siunitx}
\usepackage{qting}
\usepackage{array}

\newcommand{\rev}{\textcolor{black}}

\title{Lower-limb kinematics and kinetics during continuously varying human locomotion
}

\author[1,$\dag$]{Emma Reznick}
\author[2,3,4$\dag$]{Kyle R. Embry}
\author[5]{Ross Neuman}
\author[1,6]{Edgar Bol\'ivar-Nieto}
\author[5]{Nicholas P. Fey}
\author[1,6,*]{Robert D. Gregg}

\affil[1]{University of Michigan, Robotics Institute, Ann Arbor, MI, 48109, USA}
\affil[2]{University of Texas at Dallas, Department of Mechanical Engineering, Richardson, TX, 75080, USA}
\affil[3]{Shirley Ryan AbilityLab, Center for Bionic Medicine, Chicago, IL, 60611, USA}
\affil[4]{Department of Physical Medicine and Rehabilitation,
Northwestern University, Chicago, IL, 60611, USA}
\affil[5]{University of Texas at Austin, Department of Mechanical Engineering, Austin, TX, 78712, USA}
\affil[6]{University of Michigan, Department of Electrical Engineering and Computer Science, Ann Arbor, MI, 48109, USA}

\affil[*]{corresponding author: Robert D. Gregg (\texttt{rdgregg@umich.edu})}

\affil[$\dag$]{these authors contributed equally to this work}

\begin{abstract} 
Human locomotion involves continuously variable activities including walking, running, and stair climbing over a range of speeds and inclinations as well as sit-stand, walk-run, and walk-stairs transitions. Understanding the kinematics and kinetics of the lower limbs during continuously 
varying locomotion is fundamental to developing robotic prostheses and exoskeletons that assist in community ambulation. However, available datasets on human locomotion neglect transitions between activities and/or continuous variations in speed and inclination during these activities. This data paper reports a new dataset that includes the lower-limb kinematics and kinetics of ten able-bodied participants walking at multiple inclines ($\pm$ \SIlist{0; 5; 10}{\degree}) and speeds (\SIlist{0.8; 1; 1.2}{\meter\per\second}), running at multiple speeds (\SIlist{1.8; 2; 2.2; 2.4}{\meter\per\second}), walking and running with constant acceleration ($\pm$ \SIlist{0.2; 0.5}{\meter\per\second\squared}), and stair ascent/descent with multiple stair inclines (\SIlist{20; 25; 30; 35}{\degree}). This dataset also includes sit-stand transitions, walk-run transitions, and walk-stairs transitions. Data were recorded by a Vicon motion capture system and, for applicable tasks, a Bertec instrumented treadmill.
\end{abstract}

\begin{document}

\flushbottom
\maketitle

\thispagestyle{empty}

\section*{Background \& Summary}
To address limitations in amputee locomotion \cite{Hood2020}, robotic prosthetic legs are being developed with specifications for design and control based on able-bodied human biomechanics data \cite{Azocar2018,EleryTRO2020,Lenzi2018,Lawson2014}. Although wearable robotic devices can have different goals (e.g., reducing energetic cost \cite{zhang2017human,lee2018autonomous}), able-bodied data are often used as reference trajectories in the control system \cite{Quintero2018,Embry2020,Best:IROS2021,kang2019effect,wang2018comfort} to restore normative biomechanics in impaired individuals, such as amputees, who otherwise could not walk normally. Most studies of able-bodied human locomotion report lower-limb kinematics and kinetics during a limited set of steady-state tasks, e.g., walking \cite{Winter1983, Embry2018}, running \cite{Novacheck1998}, or stair ascent and descent \cite{Riener2002}, with only a few discrete samples of speed and/or incline for each task \cite{Embry2018}. State-of-art control systems for robotic prosthetic legs are similarly limited to a small set of steady-state locomotion tasks, using finite state machines to control instantaneous transitions between them (risking classification errors and jerky motion at transition points \cite{tucker2015control}). However, real-life human locomotion is far from steady state, involving intermittent bouts of walking, stopping, sitting, standing, and stair climbing \cite{Orendurff2008}. In fact, \SI{75}{\percent} of all walking bouts are less than 40 steps in a row \cite{Orendurff2008_boutDuration}. Non-steady conditions including transitions between locomotion modes and continuous variations of slopes and speeds are critical to modeling human locomotion and designing agile robotic prostheses.

Although the biomechanics of able-bodied walking \cite{Winter1983}, running \cite{Novacheck1998}, sit to stand \cite{Nuzik1986}, and stair climbing \cite{Riener2002} have been well-documented as independent locomotion tasks, it is difficult to combine datasets due to differences in the measurements, methods, and participants. Brantley et al. \cite{Brantley2018} recorded electroencephalography, lower-limb electromyography (EMG), and full body kinematics for ten participants walking on level-ground, ramps, and stairs. Embry et al. \cite{Embry2018} reported lower-body kinematics, kinetics, and EMG activity for ten able-bodied participants while walking at multiple speeds and slopes. Schreiber and Moissenet \cite{Schreiber2019} reported whole body kinematics, kinetics, and lower-body EMG information for level-ground walking at different speeds. Lencioni et al. \cite{Lencioni2019} reported whole body kinematics, kinetics, and lower-body EMG information for level-ground walking and stair ascent/descent. Hu et al. \cite{Blair18} recorded lower-limb kinematics and EMG of ten able-bodied participants during free transitions between sitting, standing, level-ground walking, ramp walking, and stair climbing. Camargo et al. \cite{Young:dataset} recorded kinematics, kinetics, and EMG for 22 participants during variable-speed treadmill walking, variable-incline overground walking, variable-height stair climbing, and ramp and stair transitions. However, these studies are missing one or more of the following: joint kinetic measurements, global orientation measurements (e.g., pelvic tilt), enrollment of older participants, a primary activity of daily living (e.g., sit-to-stand), transitions between tasks, and/or continuous variations of walking speed \emph{and} incline (including their combinations).

\begin{figure}[ht]
	\centering
	\includegraphics[]{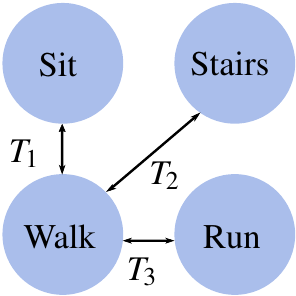}
	\caption{Locomotion modes and practical transitions (T1 through T3) considered in this study. `Walk' and `Run' occur on a flat surface, whereas `Stairs' occurs on a staircase. All modes except `Sit' are continuously parameterized by speed and inclination, which are sampled in the dataset. Note a sit-stand transition corresponds to $T_1$ with zero gait speed in `Walk.'}
	\label{fig:PracticalTransitions}
\end{figure}

This data paper considers all these features in reporting the lower-limb measurements of ten able-bodied participants during sit-stand, walk-run, and walk-stairs transitions (\autoref{fig:PracticalTransitions}), walking at multiple inclines with different speeds, walking/running at constant acceleration and deceleration rates, and running at multiple speeds. The reported kinematics include pelvic tilt and hip, knee, and ankle joint angles, and joint moments are provided for the treadmill tasks. 
The dataset also includes the global position of reflective markers (hereinafter referred to as ``markers'') in 3D space and the force plate measurements when available. These measurements were used to build kinematic and kinetic models to calculate joint angles, velocities, and moments. We also include video recordings to illustrate the experiments for each of the locomotion tasks. 

One purpose for the collected dataset is to provide input data for modeling human kinematics over different locomotion tasks. During a steady gait cycle, each joint angle can be assumed to be a periodic signal and thus can be modeled as a Fourier series depending on time or a global measurement \cite{Quintero2018}. The changes in knee and ankle kinematics due to different walking speeds and inclines can be modeled as a weighted sum of multiple continuous basis functions, e.g., Fourier series \cite{Embry2018}. This allows the model to continuously interpolate walking kinematics over continuously varying speeds and inclines, giving robotic prosthetic legs more adaptability than switching between a limited set of discrete inclines or speeds. Continuous transitions between sitting, standing, walking, stair climbing, and running can also be modeled for more natural control of non-steady activities. These kinematic models can be trained based on across-participant averages from the presented dataset to generate baseline control strategies for powered prostheses \cite{Quintero2018,Embry2020,Best:IROS2021} and exoskeletons \cite{kang2019effect,wang2018comfort}. Because these models continuously connect a range of tasks, they can be efficiently individualized by heavily weighting one participant-specific task (e.g., level-ground walking) amongst across-participant averages for all other tasks \cite{Reznick2020}.

\section*{Methods}
This section lays out the procedure from obtaining participant consent to acquiring and processing the data for publication. Each section proceeds step-by-step through the procedure and details the tasks within each ambulation mode. The variety of speeds and/or inclines within each mode entail the continuous variation of the dataset.    


\subsection*{Participants}
The study protocol was approved by the Institutional Review Boards at the University of Texas at Dallas and the University of Michigan. This dataset was acquired from ten healthy participants (5 female), aged 20-60 years (30.4 $\pm$ 14.9), weighing 74.6 $\pm$ \SI{9.7}{kg}, and with an average height of 1.73 $\pm$ \SI{.94}{\meter}. To be included, the participants were between 19 and 65 years old, self-reported the ability to walk over uneven ground with ease, and had no joint problems in the lower extremities or neuromuscular disorders or diseases that would impair their ability to walk.

\subsection*{Instrumentation and Participant Preparation}
Data were acquired in the University of Texas Southwestern Medical Center motion capture laboratory in Dallas, TX, operated by University of Texas at Dallas. This laboratory is equipped with a motion capture system, instrumented treadmill, and a 4-step adjustable stair set in order to record the kinematics and kinetics of a wide range of tasks, detailed below. Synchronous data acquisition was managed by the Vicon motion capture system.

\subsubsection*{Participant Preparation} \label{participant_prep} 
\begin{figure}[ht]
\centering
\includegraphics[width=\textwidth, trim = 0 0 0 0, clip]{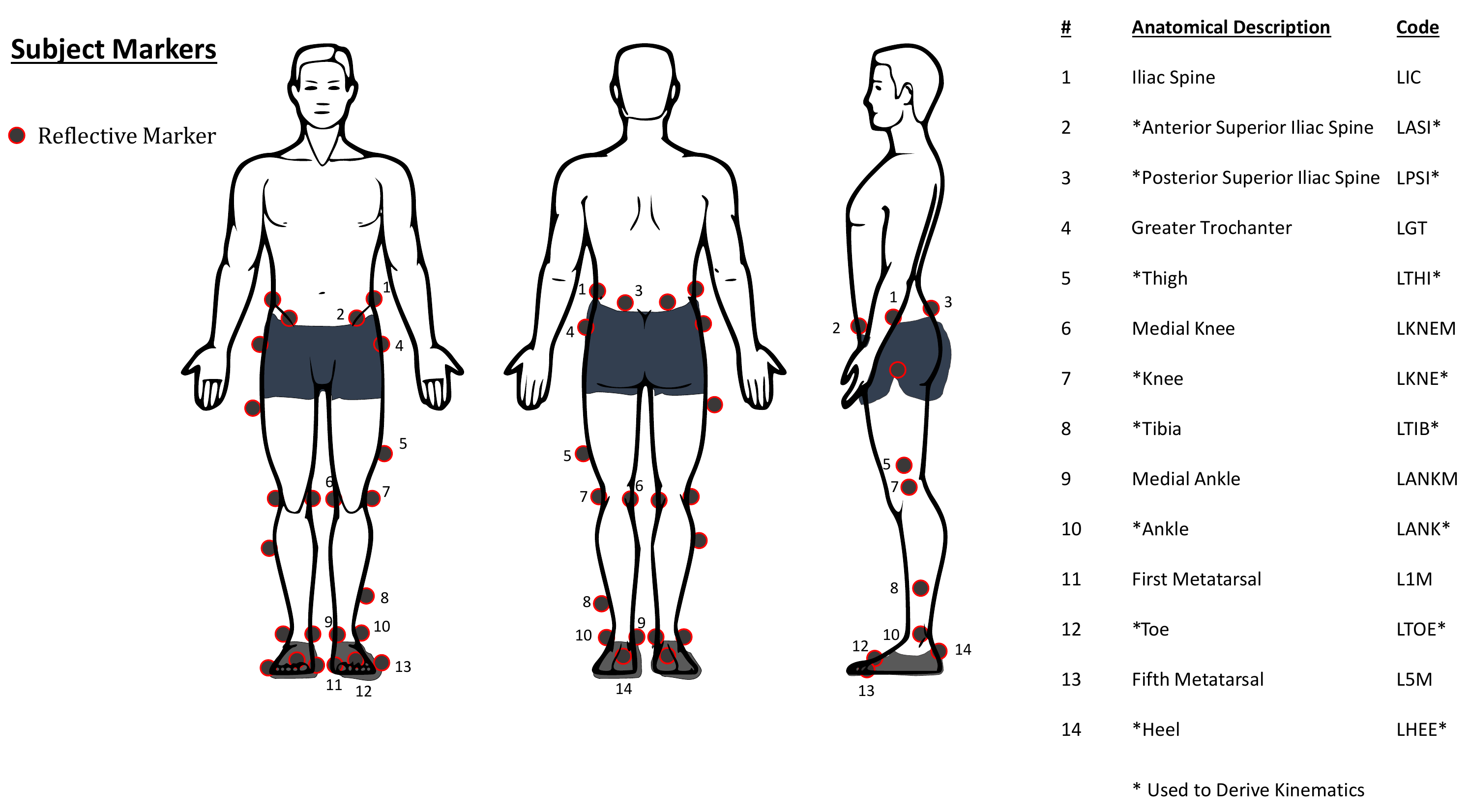}
\caption{Marker Set: Markers were placed on left and right limbs symmetrically, left side markers shown. Markers noted with a `*' are used in the conventional gait model. For full description of marker locations, see Nexus 2 user guide \cite{nexusUserGuideLowerBody}.}
\label{fig:Marker Set}
\end{figure}

After obtaining informed consent and briefing participants about all trials that would be performed, participants were equipped with motion capture markers. Participants were instructed to wear tight-fitting clothing, \rev{and cohesive wraps were used when necessary to ensure clothing stayed tight to the skin, particularly for two participants (AB06 and AB10)}. Markers were placed at every location necessary for the Vicon Plug-in Gait lower body model, as well as a custom set of additional markers used to improve the robustness of segment identification. These additional markers provided a source for filling either a rigid body or pattern, depending on the location of the marker, during events where a required marker for Plug-in Gait was not visible for several frames.
\rev{Note that some of the additional markers were placed close together on 3D printed cluster plates held to the leg with cohesive tape, but these redundant markers are left out of the dataset due to possible motion artifacts.} Please see \autoref{fig:Marker Set} for the names of all markers and a description of which markers are used for the conventional gait model. \rev{The Plug-in Gait markers were almost always placed on skin or tight-fitting clothes (e.g., yoga pants), with the exception of the thigh markers for only two participants (AB06 and AB10) which were subsequently constrained to the rigid body during post-processing.}

\subsubsection*{Treadmill} All flat-ground walking and running tests were conducted at variable speeds and inclines on a Bertec instrumented split-belt treadmill (Bertec Corporation, Columbus, OH). This treadmill has embedded force plates under each belt to acquire kinetics independently for each leg at 1000 Hz (and downsampled to 100 Hz for publication). The treadmill was remotely controlled by a custom MATLAB code to change the speed and acceleration over randomized tasks at one incline, details discussed later.

\subsubsection*{Motion Capture} A 10-camera Vicon T40 motion capture system (Vicon, Oxford, UK) was used to record the 3D positions of all markers attached to participants at 100 Hz. We utilized Vicon's proprietary Dynamic Plug-in Gait Model to calculate joint angles from markers positions, where angular conventions are defined in the Nexus 2 user guide\cite{nexusUserGuide}. In trials using the treadmill forceplates, Plug-in Gait was also used to approximate joint forces, moments, and powers. The Plug-in Gait software calculates inverse dynamics to derive joint kinetics from force plate kinetics and motion capture kinematics using the conventional gait model. Further information on the Plug-in Gait kinetic modeling is available in the Nexus 2 user guide \cite{nexusUserGuide}. An open-source implementation of the conventional gait model is introduced in Lebourd et. al \cite{Leboeuf2019}, which closely matches Plug-in Gait results. In our dataset, each joint has a consistent sign convention for its angles and moments, where ankle dorsiflexion, knee flexion, and hip flexion are defined in the positive directions.

\subsubsection*{Stair Details} A 4-Step Adjustable Stair Set (Staging Dimensions, Inc., New Castle, Delaware) with no handrails was attached to a platform with variable height legs. The height of the platform was adjusted to change the inclination of the stairs. This allows for stair heights that closely resemble the 2010 ADA accessibility standards of 4-7in. The inclines (\SIlist{20; 25; 30; 35}{\degree}) correspond with riser heights of \SIlist{3.81; 4.72; 5.75; 6.38}{in}. Due to lack of force plates in the stairs, kinetic data was not recorded for any stair walking. 


\subsection*{Experimental Procedure}
This study captures the kinematics, force plate and joint kinetics of walking and running on a flat surface, kinematics and force plate kinetics of sit-to-stand, and kinematics of stair ascent and descent. The acquisition of each of these ambulation modes consists of a continuous range of tasks that a participant may encounter in day-to-day life, including different inclinations and speeds and task transitions (\autoref{fig:PracticalTransitions}). While the experiment was designed to randomize different tasks to minimize the effect of fatigue, the progression of ambulation modes (described later) remained constant for ease of acquisition. 

We performed a statistical analysis before conducting this experiment to ensure reasonable confidence that we enrolled enough participants, recorded enough strides of each task, and that our sample mean kinematics would be reasonably close to the population mean. Our metric for success was that based on the 80\% confidence interval of the standard deviation of previous experiments, the intra-participant mean trajectory should be within five degrees of the true intra-participant trajectory at all points in time, and similarly that the inter-participant mean is within five degrees of the population inter-participant mean \cite{Embry2018}. The number of strides and participants necessary to meet this condition was found by the formula $n = \left( Z_{\alpha / 2} \sigma/E \right)^2,$ where $n$ is the necessary number of samples, $Z_{\alpha / 2} = 1.645$ is the $\alpha=0.10$ z-score for a two-tailed distribution,  $\sigma$ is the population standard deviation of the data, and $E$ is the maximum difference between the population mean and $n$-sample mean, in our case 5 degrees. The population standard deviation $\sigma$ is unknown, so we replace this value with the upper and lower 80\% confidence interval of the standard deviation, $s$, of a previous, similar experiment \cite{Embry2018} to produce a range of $n$ values (see \autoref{tab:sampleSize}). We selected practical values of $n$ within this range that are consistent with many other human motion capture studies.

\subsubsection*{Participant Marker Placement and Calibration} 
After obtaining informed consent and preparing the participant's marker set (\autoref{fig:Marker Set}), the participant was instructed to stand still for a static calibration of the marker set, and then walk for 10-15 seconds on the treadmill at \SI{1.0}{\meter\per\second} while we recorded their joint kinematics. Any unexpected irregularities in able-bodied gait were addressed at this point before proceeding. Please see the Technical Validation section for details. 

\subsubsection*{Walking} 
\begin{figure*}
\centering
\includegraphics{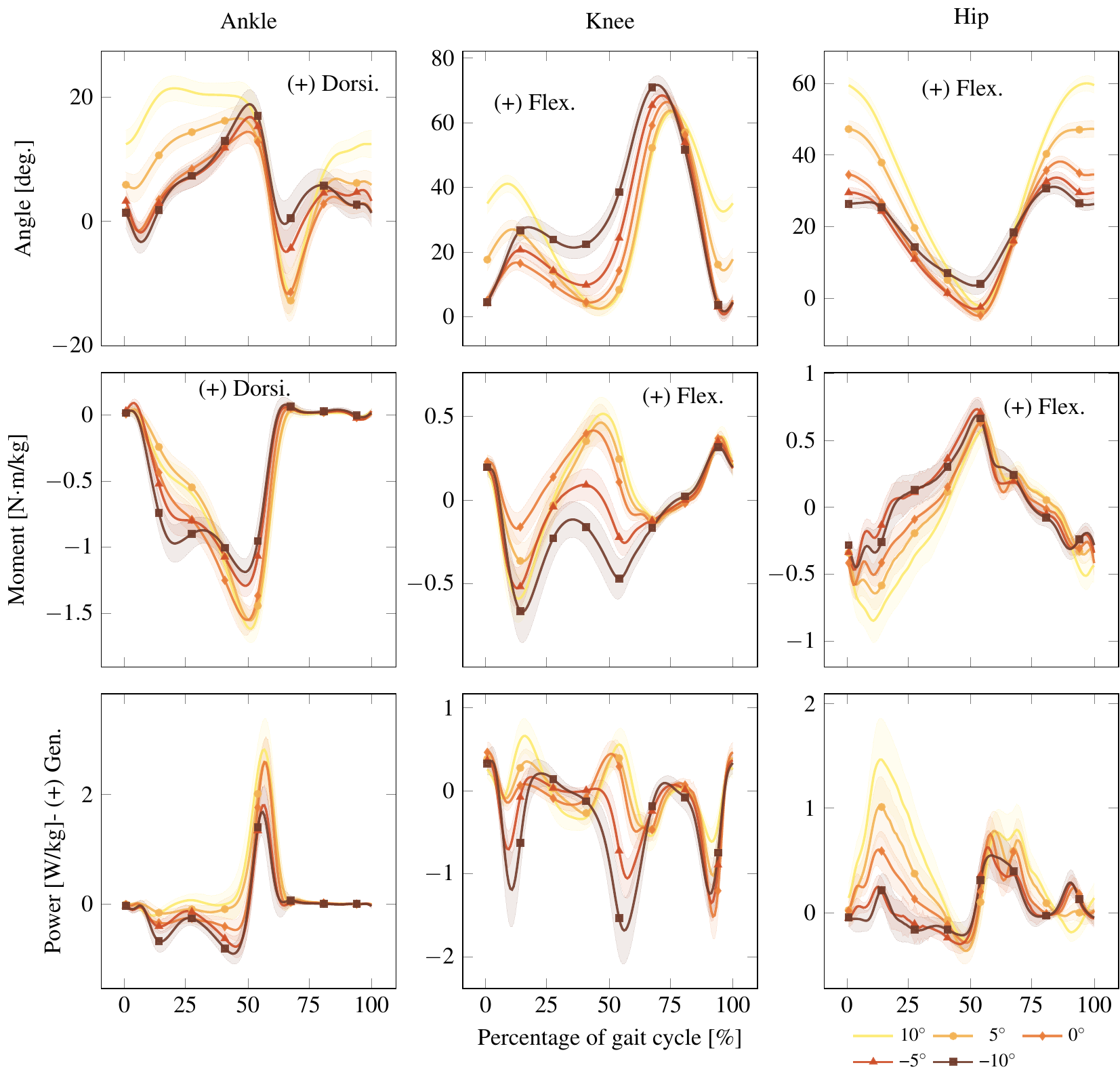}
\caption{Inter-participant average walking kinematics and kinetics for all inclines at \SI{1.0}{\meter\per\second}. Inclines reported in degrees. Foot contact corresponds to \SI{0}{\percent} of the gait cycle. Solid lines and shaded regions represent the average trajectory and its variation within one standard deviation, respectively. Positive or negative normalized power denotes generation (Gen.) or absorption (Abs.) of mechanical power, respectively. These plots correspond to {\tt Walk} in Normalized.mat, as detailed in the README.}
\label{fig:walk}
\end{figure*}

A MATLAB program was used to remotely control the treadmill through a protocol that randomized the order of walking speed tasks as well as two acceleration tasks. Walking trials progressed through a range of nominal walking speeds (\SIlist{0.8; 1.0; 1.2}{\meter\per\second}) over the course of each capture. 
In an effort to increase similarity between participant joint angles, the nominal walking speeds were then normalized with respect to the participant's leg length, using the formula $v_{\mathrm{norm}} = v_{\mathrm{orig}}/\sqrt{gl_0}$, where $g = 9.81$ is the gravity constant and $l_0$ is leg length \cite{Hof1996}. Different acceleration and deceleration rates were also tested ($\pm$ \SIlist{0.1; 0.2; 0.5}{\meter\per\second^2}), where \SI{0.1}{\meter\per\second^2} was used to transition between different walking speeds, and \SI{0.2}{\meter\per\second^2} and \SI{0.5}{\meter\per\second^2} were tested individually by accelerating from rest to \SI{1.2}{\meter\per\second}, holding for 5 seconds, and decelerating back to rest. These procedures were conducted at multiple inclines covering and slightly exceeding the range of ADA-compliant ramps ($\pm$ \SIlist{0; 5; 10}{\degree}), the order of which was also randomized to minimize the effect of fatigue. These walking speeds and inclines match some of our previously released datasets, which may be useful for comparison \cite{embryDataport, macalusoDataport, eleryDataport}. Walking kinematics and kinetics over each incline are shown in \autoref{fig:walk}. Each data capture was conducted at a fixed incline with the treadmill incline feature clamped to prevent surface movement while walking. A tone was played to alert the participant before any changes in speed.

\subsubsection*{Running}
\begin{figure*}
\centering
\includegraphics{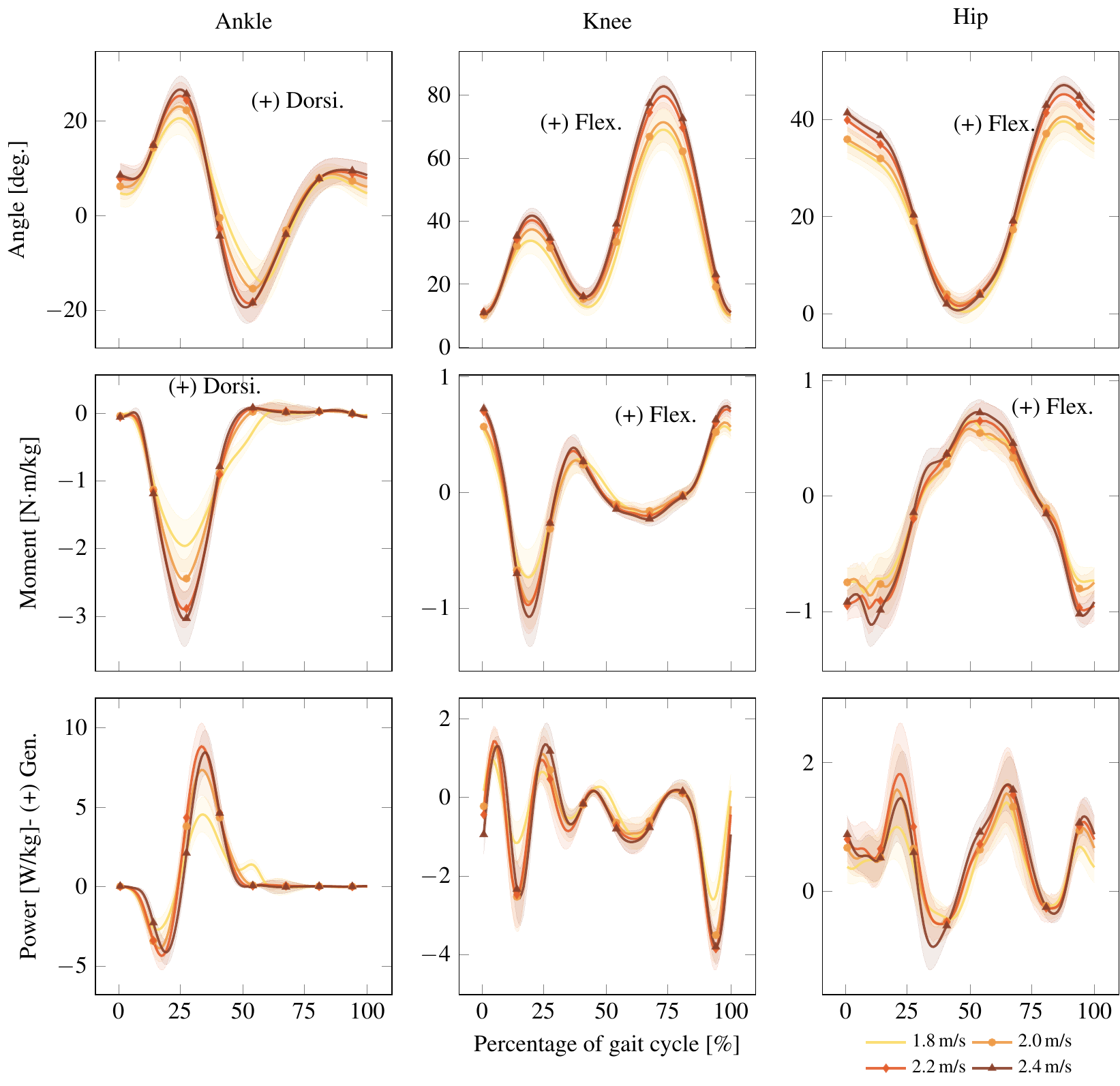}
\caption{Inter-participant average running kinematics and kinetics at all speeds. Foot contact corresponds to \SI{0}{\percent} of the gait cycle. Solid lines and shaded regions represent the average trajectory and its variation within one standard deviation, respectively. Positive or negative normalized power denotes generation (Gen.) or absorption (Abs.) of mechanical power, respectively. These plots correspond to {\tt Run} in Normalized.mat, as detailed in the README.}
\label{fig:run}
\end{figure*}

Running trials were collected on the treadmill in a randomized order of speeds over level ground (\SIlist{1.8; 2.0; 2.2; 2.4}{\meter\per\second}), with speeds normalized by leg length consistent with walking trials \cite{Hof1996}. Data was collected for 30 seconds at each speed, resulting in the inter-participant average kinematics and kinetics in \autoref{fig:run}. Walk-run transitions were separately captured and remotely controlled by accelerating from rest to \SI{2.2}{\meter\per\second} at different rates (\SIlist{0.2; 0.5}{\meter\per\second^2}), holding that speed for 10 seconds, and decelerating at the same rate back to rest. It should be noted that AB04 and AB10 opted out of performing some running trials.

\subsubsection*{Sit-to-Stand} 
\begin{figure*}
\centering
\includegraphics{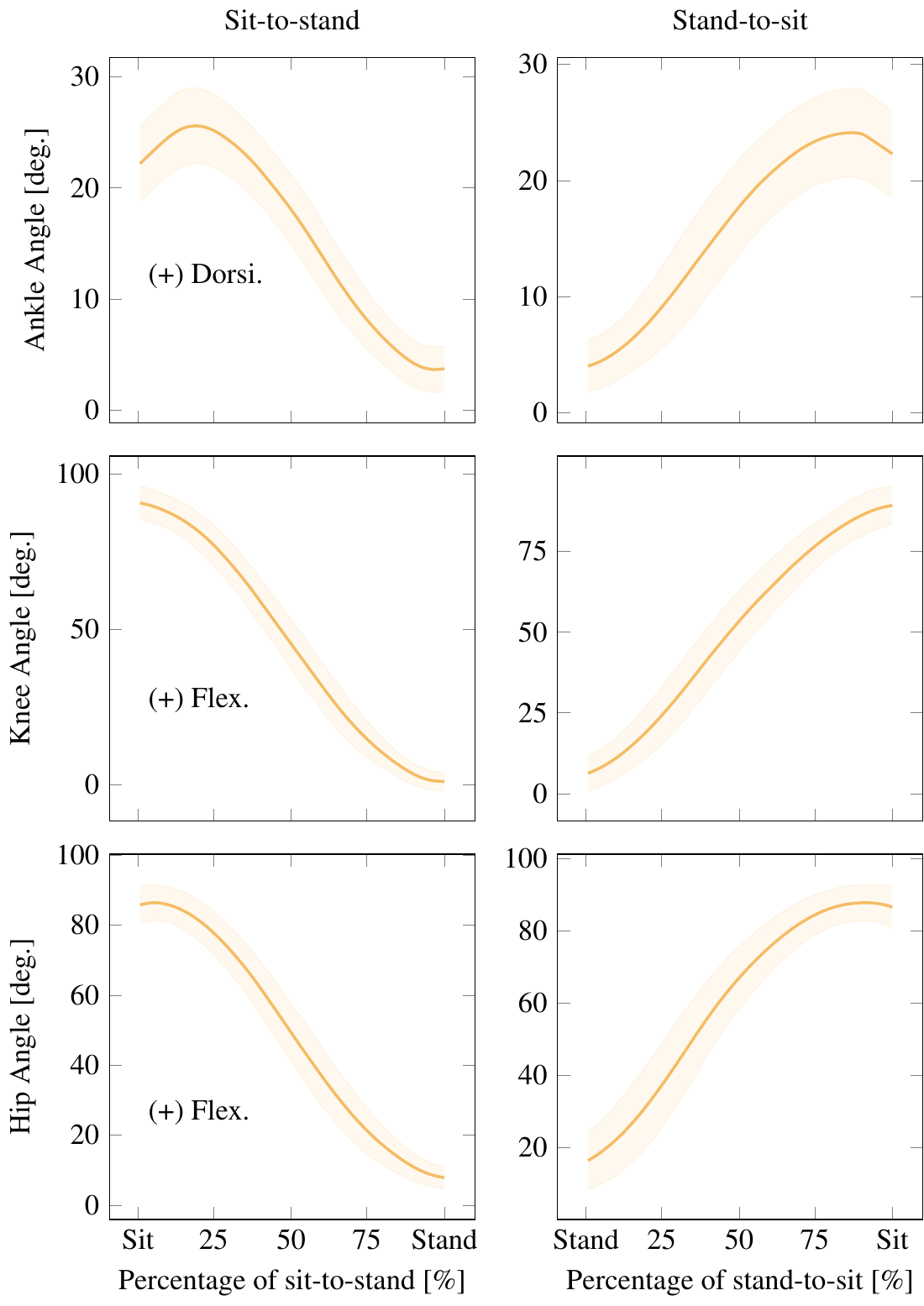}
\caption{Inter-participant average sit-to-stand kinematics. Solid lines and shaded regions represent the average trajectory and its variation within one standard deviation, respectively. These plots correspond to {\tt SitStand} in Normalized.mat, as detailed in the README.}
\label{fig:sit2stand}
\end{figure*}

Sit-to-stand transitions were collected by instructing the participant to sit on a backless stool placed on one of the belts of the treadmill while their feet rested on the other belt. Force plates beneath each belt recorded the ground reaction forces through the stool and the participant's feet throughout the transition, allowing researchers to study the weight transition to and from the feet throughout the sit-to-stand and stand-to-sit transitions. Six trials were recorded of the participant rising from the chair, standing at rest for a moment, and then sitting. Participants were instructed not to use their hands to assist their transitions. The inter-participant average kinematics can be seen in \autoref{fig:sit2stand}.

\subsubsection*{Stairs}
\begin{figure*}
\centering
\includegraphics{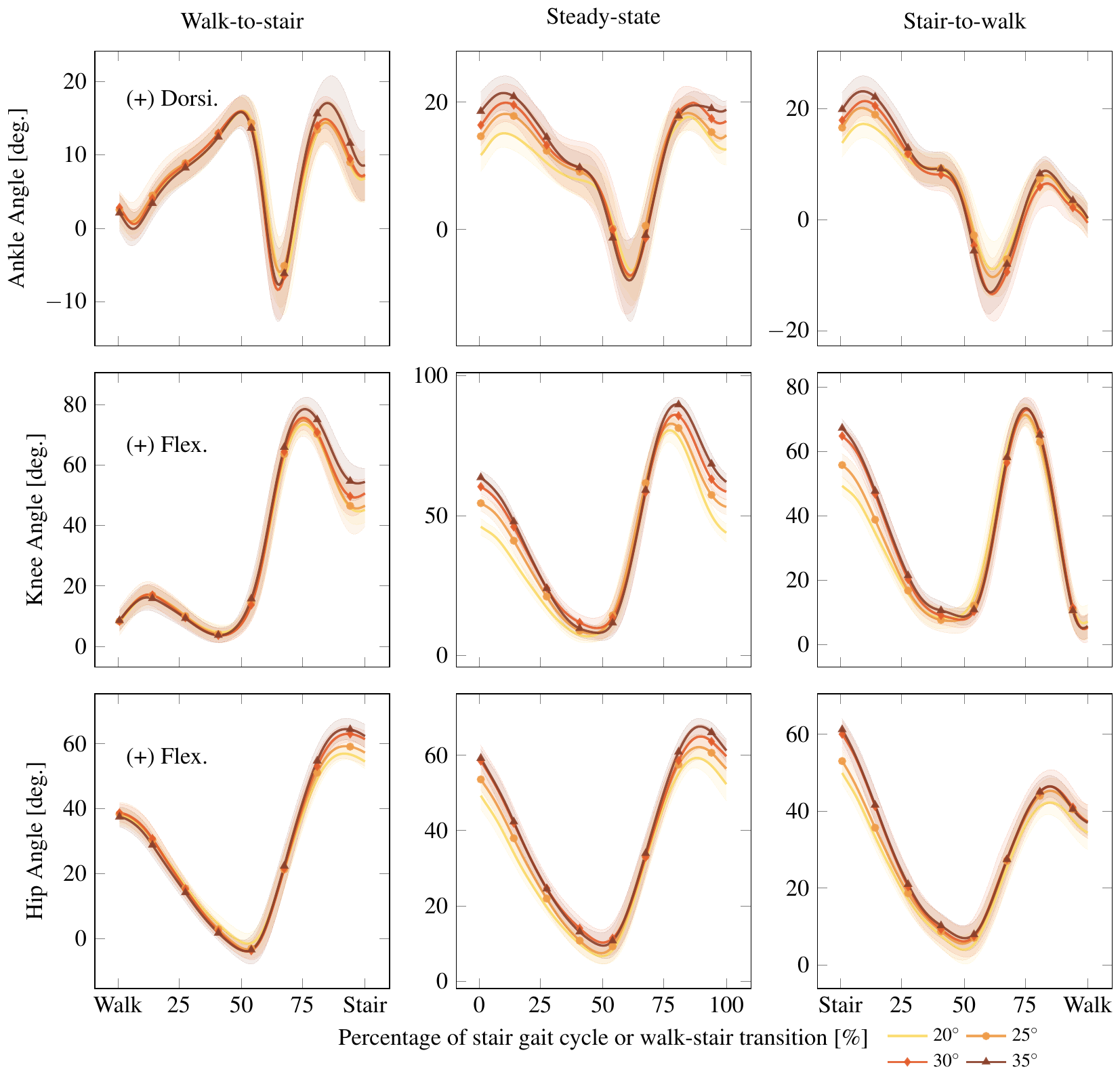}
\caption{Stair ascent steady-state and transition kinematics. Inter-participant average kinematics for steady-state stair ascent and the transitions walk-to-stair-ascent and stair-ascent-to-walk. Stride 3 is plotted for the steady-state case because it is the most periodic stride. Stair inclines are reported in degrees. Solid lines and shaded regions represent the average trajectory and its variation within one standard deviation, respectively. These columns respectively correspond to the ascent cases of {\tt w2s}, {\tt s3}, and {\tt s2w} under {\tt Stair} in Normalized.mat, as detailed in the README.}
\label{fig:stair-ascent}
\end{figure*}

\begin{figure*}
\centering
\includegraphics{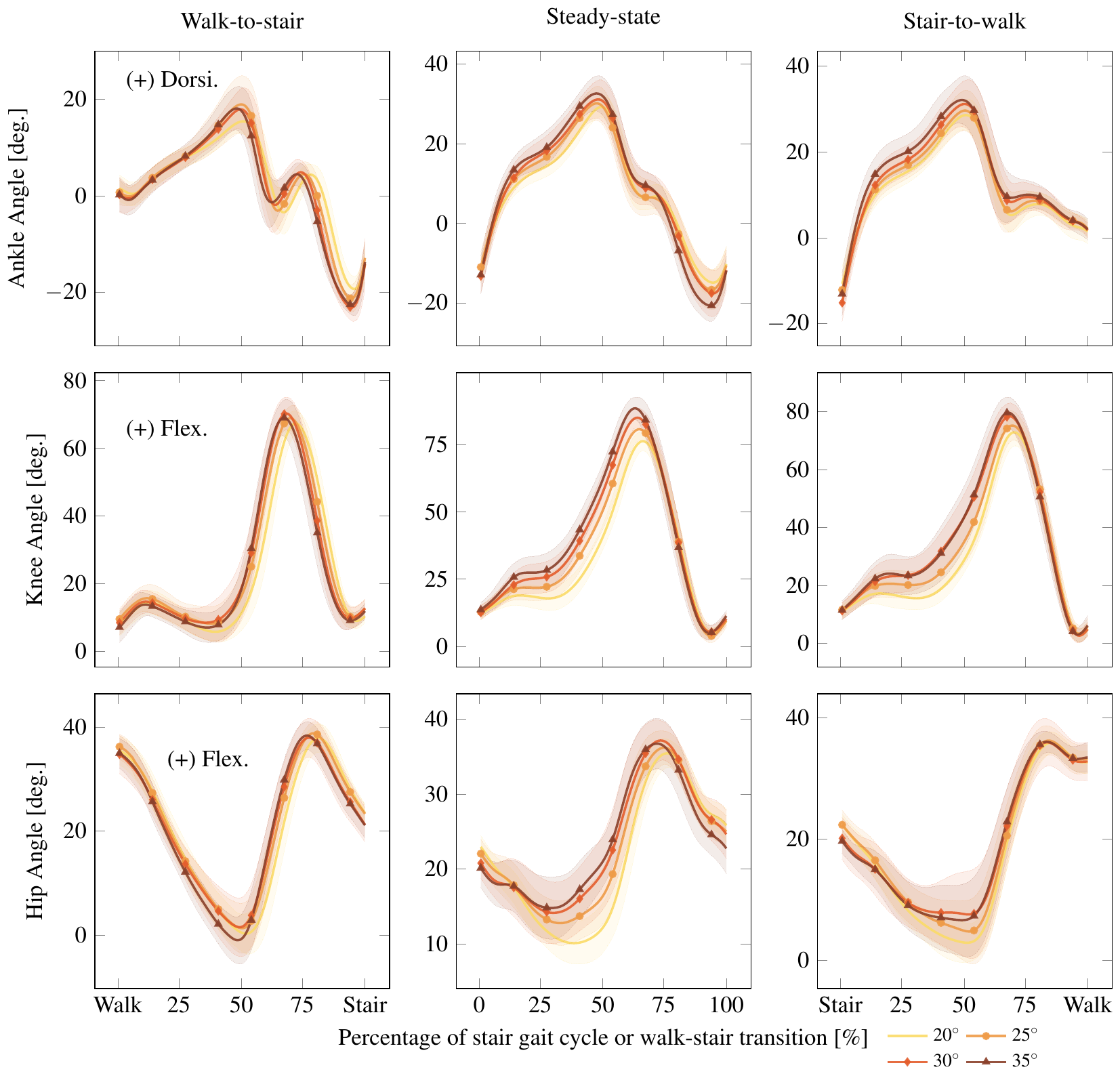}
\caption{Stair descent steady-state and transition kinematics. Inter-participant average kinematics for steady-state stair descent and the transitions walk-to-stair-descent and stair-descent-to-walk. Stride 3 is plotted for the steady-state case because it is the most periodic stride. Stair inclines are reported in degrees. Solid lines and shaded regions represent the average trajectory and its variation within one standard deviation, respectively. These columns respectively correspond to the descent cases of {\tt w2s}, {\tt s3}, and {\tt s2w} under {\tt Stair} in Normalized.mat, as detailed in the README.}
\label{fig:stair-descent}
\end{figure*}

Stair trials were conducted over four inclinations of stairs (\SIlist{20; 25; 30; 35}{\degree}). The participant began 6 ft from the base of the staircase, approached the stairs at a self-selected walking speed, ascended the stairs, and walked to the end of the platform. At this point, the capture was ended and the participant was instructed to turn around. This procedure recorded one walk-to-stair and stair-to-walk transition per capture. The participant then began the descent trial from rest at the end of the platform, descended to the bottom of the stairs, and continued to the demarcated starting line. At least five ascent and descent trials were conducted at each incline, with additional trials being added at the recorder's discretion. The inter-participant average kinematics of steady-state and transitional strides are shown for stair ascent in \autoref{fig:stair-ascent} and stair descent in \autoref{fig:stair-descent}.


\subsection*{Motion Capture Post-Processing} Post processing in Vicon Nexus consisted of rigid body fills, filtering, and a Plug-in Gait model that uses the conventional gait model and inverse dynamics to calculate joint moments. After labeling the markers to match the marker set in \autoref{fig:Marker Set}, a rigid body or pattern fill addressed small gaps for each leg segment ($L_{\mathrm{thigh}}$, $L_{\mathrm{shank}}$, $R_{\mathrm{thigh}}$, $R_{\mathrm{shank}}$). Gaps in the data were caused by visual occlusion of the markers. The marker trajectories were then filtered with a $4^{th}$-order Butterworth Low-Pass filter with a 6Hz cutoff \cite{winter2009biomechanics} and a Woltring filter with a smoothing parameter of 20 . Finally, the Plug-in Gait model calculated joint kinematics and kinetics (when force plate data was available), and the data was checked for anomalous motion. If anomalies were found, the motion was corrected at the marker level and the model was run again without filters to prevent over-filtering.

\subsection*{Data Processing} A custom MATLAB pipeline was implemented to scan through all kinematic trajectories for a given participant, parse trials into tasks, parse tasks into strides, time-normalize strides, and compile into a unified MATLAB structure. For walking trials, the treadmill controller was programmed to output a command log after each trial detailing the sequence and timing of the various commands, which provided the necessary information to separate a trial into its constituent tasks. A stride was defined as heel strike to subsequent heel strike for a given leg. For all tasks performed on the treadmill, heel strikes were detected using forceplate measurements using software provided by Vicon \cite{detectEvents}. For stair trials, heel strikes were manually labeled in Vicon Nexus. Heel strikes also determined the periods for stance and swing phases, which are given in the \texttt{events} struct of the dataset. Each stride was considered a repeated measure of the given task. For sit-to-stand trials, tasks were separated algorithmically by tracking the progression of sagittal plane knee angles, see \autoref{fig:sit2stand}. After heel strike detection, strides were linearly interpolated to 150 data points. Since all participants in this dataset were able-bodied and assumed to present symmetric kinematics, all joint kinematic fields include data from both the right and left legs without specific side labelling. Once the strides of a given task were compiled, unsatisfactory examples were removed based on several criteria: strides containing true zeros (i.e., gaps existed in data); strides with outliers in the maximum value of the first derivative (i.e., large discontinuities); steady-state stair strides with large differences between starting and end points (i.e., non-periodicity); and strides whose mean values were outliers (offset errors). Outliers were defined as values more than 3 scaled median absolute deviations from the median. Additional removal of fragmented strides (containing a task change) ensured veracity of the data. This strict judgment rejected 8.3\% of treadmill strides (Walk, Run, Walk-to-Run) and 7.2\% of Stair strides across all participants, leaving a median of 2006 strides for each participant across all modes and tasks. No sit-to-stand trials were rejected.
   


\section*{Data Records}
This 10-person able-bodied dataset can be accessed from the Figshare data repository\cite{figshare} in the form of two MATLAB structures: Streaming.mat is the continuous data from each trial (\autoref{Tab:raw}) and Normalized.mat contains the same data parsed and normalized by stride (\autoref{Tab:Normalized}). Further documentation for the dataset can be found in the corresponding README file.

\section*{Technical Validation}
\label{s:techvalid}

The manufacturer's standard procedure for capture volume calibration was performed before each session of experiments, and was repeated as needed if a camera was disturbed for any reason. This procedure entailed calibrating the cameras within the capture volume, leveling the treadmill, and setting the volume origin. Similarly, the force plates in the Bertec treadmill were both hardware zeroed, and software zeroed through the Nexus motion capture system. This zeroing procedure was repeated each time the treadmill inclination changed, to account for the loading conditions at the new incline. Finally, a custom procedure was implemented to calibrate the placement of motion capture markers attached to participants. First, all markers were placed against the bony landmarks described by the Plug-in Gait marker procedure. Then, participants were asked to walk on the treadmill at zero degree incline for at least ten strides and recorded by motion capture. This recording was manually checked for good symmetry between left and right side joints and that the range of motion of each joint was similar to what was expected for able-bodied participants. If it was determined that the recorded symmetry or range of motion was unnatural for the participant, one or more markers would be moved slightly, always less than one centimeter, and the procedure was repeated. This procedure was repeated until joint symmetry and range of motion were deemed within expected ranges by two researchers to prevent bias. Note that this procedure would need to be modified for patient populations with different expected range of motion or gait asymmetry.



\section*{Usage Notes}


Our dataset includes a MATLAB script entitled \texttt{exampleUseScript.m}, which gives an example of how to quickly access different types of data within the MATLAB structure. In this example script, we have plotted the sagittal-plane thigh kinematics and force plate vertical load for AB04 walking for one minute at \SI{1.0}{\meter\per\second} at \SI{-10}{\degree} slope. All 59 strides performed by this participant are superimposed in this plot. Please see \texttt{exampleUseScript.m} for full explanation on how to access this data.

\section*{Code Availability}
The dataset and the code to post-process the data and control the Bertec treadmill can be accessed though MATLAB and is documented with a README file describing the data hierarchy. The MATLAB code that remotely controls the Bertec treadmill running a designed protocol is available with the dataset. This code is documented with its own README file. 

\bibliography{R01}

\section*{Acknowledgements}
This work was supported by the National Institute of Child Health \& Human Development of the NIH under Award Number R01HD094772. The content is solely the responsibility of the authors and does not necessarily represent the official views of the NIH. Robert D. Gregg, IV, Ph.D., holds a Career Award at the Scientific Interface from the Burroughs Wellcome Fund.

The authors would like to thank Lizbeth Zamora for her help with data acquisition and data post-processing, Rebecca Macaluso for her help adapting the treadmill remote control, and Shihao Cheng, Vamsi Peddinti, Erica Santos, and Kevin Best for their help debugging the dataset.

\section*{Author contributions}
Emma Reznick planned and performed the experiment, post-processed and debugged the data, and is a co-first author of this paper. \vspace{5mm}

\noindent Kyle R. Embry planned and performed the experiment, post-processed and debugged the data, and is a co-first author of this paper.\vspace{5mm}

\noindent Ross Neuman planned and performed the experiment, wrote the code for task separation and stride normalization, and contributed to the paper.\vspace{5mm} 

\noindent Edgar Bol\'ivar assisted with participant recruitment, debugging of the dataset, and contributed significantly to writing and editing the paper.\vspace{5mm}

\noindent Nicholas Fey contributed to experimental planning as well as co-supervision of data acquisition, processing and interpretation. Dr. Fey was a senior personnel of the grant funding this work.\vspace{5mm}

\noindent Robert D.~Gregg managed the study design and execution, data processing/debugging, and contributed to paper writing and editing. Dr.~Gregg was the PI of the grant funding this work.

\section*{Competing interests}

The authors declare no competing interests.

\section*{Figures \& Tables}

\begin{table}[ht]
\centering
\begin{tabular}{l|l|l|l|l|l|}
\cline{2-6}
              & Lower $s$ & Upper $s$ & Lower $n$ & Upper $n$ & Selected $n$ \\ \hline
intra-participant & 6.95       & 8.22       & 2.28       & 4.36       & 5       \\\hline
inter-participant & 7.34       & 15.83       & 3.480       & 16.17       & 10        \\\hline
        
\end{tabular}
\caption{Sample Size Calculations}
\label{tab:sampleSize}
\end{table}

\begin{table}[p]\centering
\caption{\textbf{Streaming.mat} - data without parsing/normalizing}
\begin{tabular}{m{2.2cm} m{2cm} m{2cm} m{8cm} }
\hline
\textbf{Field within structure} & \textbf{Units} & \textbf{Sampling rate} & \textbf{Contents} \\ \hline
\texttt{marker} & (m) & 100 Hz & 
Position in the global coordinate frame of the markers defined in 
\autoref{fig:Marker Set}.\newline Array Format: (total frames x 4) \newline First Dimension: Frames in trial \newline Second Dimension: x/y/z/e location in global space, e is whether the marker exists \\ \hline

\texttt{jointAngle} & (deg) & 100 Hz & Pelvic tilt, hip, knee, and ankle angles as defined in the Vicon's Plug-in Gait model \cite{Leboeuf2019}. \newline Array Format: (total frames x 3)\newline First Dimension: Frames in trial\newline Second Dimension: x/y/z rotation in local space \\ \hline

\texttt{jointForce} & (N/kg) & 100 Hz & Force vectors acting on the hip, knee, and ankle joints, given in the more distal segment's frame of reference. \newline Array Format: (total frames x 3)\newline First Dimension: Frames in trial\newline Second Dimension: x/y/z joint force \\ \hline

\texttt{jointMoment} & (N.m/kg) & 100 Hz & Pelvic tilt, hip, knee, and ankle moments normalized by the participant's mass as defined in the Vicon's Plug-in Gait model \cite{Leboeuf2019}. \newline Array Format: (total frames x 3)\newline First Dimension: Frames in trial\newline Second Dimension: x/y/z joint moments \\ \hline

\texttt{jointPower} & (W/kg) & 100 Hz & Estimated power at each joint normalized by the participant's mass. It results from the multiplication of \texttt{jointMoment} and the estimated joint velocity. \newline Array Format: (total frames x 3)\newline First Dimension: Frames in trial\newline Second Dimension: x/y/z joint power \\ \hline

\texttt{forceplates} & Force: (N)\newline Moment: (N.m)\newline COP: (m) & 1000 Hz & Resulting forces and moments in a 3D-frame with origin at the center of pressure (COP). The force plates are embedded in the two belts of the instrumented treadmill.\newline Array Format: (total frames x 3)\newline First Dimension: Frames in trial\newline Second Dimension: x/y/z of variable in global space\\ \hline

\texttt{events} & LHS: (frame)\newline RHS: (frame)\newline StrideTime: (frame)\newline VelProf: (m/s) & N/A & Heel Strikes used for normalization (L/R) and the duration of each stride (L/RStrideTime).\newline Array Format: (1 x HS or stride)\newline Velocity Profiles (Walk only: Cvel,Rvel,Lvel): Commanded treadmill velocity (m/s), L/R stride velocity (m/s)\newline Array Size: (1 x frame)  \\ \hline
\end{tabular}
\label{Tab:raw}
\end{table}

\begin{table}
\centering
\caption{\textbf{Normalized.mat} - data parsed and normalized by strides/cycles}
\begin{tabular}{lccl}
\hline
\textbf{Field within structure} & \textbf{Unit} & \textbf{Contents} \\ \hline

\texttt{marker} & (m) & \begin{tabular}[c]{@{}l@{}}Array Format: (150 x 4 x stride) \\ First Dimension: normalized sample gait over the stride/task (150pt) \\ Second Dimension: x/y/z/e location/existence in global space\\ Third Dimension: stride number\end{tabular} \\ \hline
\texttt{jointAngle} & (deg) & \begin{tabular}[c]{@{}l@{}}Array Format: (150 x 3 x stride)\\ First Dimension: normalized sample gait over the stride/task (150pt) \\ Second Dimension: x/y/z rotation in local space\\ Third Dimension: stride number (L/R leg have been concatenated)\end{tabular} \\ \hline
\texttt{jointForce} & (N/kg) & \begin{tabular}[c]{@{}l@{}}Array Format: (150 x 3 x stride)\\ First Dimension: normalized sampled gait over the stride/task (150pt) \\ Second Dimension: x/y/z joint force \\ Third Dimension: stride number (L/R leg have been concatenated)\end{tabular} \\ \hline
\texttt{jointMoment} & (N.m/kg) & \begin{tabular}[c]{@{}l@{}}Array Format: (150 x 3 x stride)\\ First Dimension: normalized sampled gait over the stride/task (150pt) \\ Second Dimension: x/y/z joint moment \\ Third Dimension: stride number (L/R leg have been concatenated)\end{tabular} \\ \hline
\texttt{jointPower} & (W/kg) & \begin{tabular}[c]{@{}l@{}}Array Format: (150 x 3 x stride)\\ First Dimension: normalized sampled gait over the stride/task (150pt) \\ Second Dimension: x/y/z joint power \\ Third Dimension: stride number (L/R leg have been concatenated)\end{tabular} \\ \hline
\texttt{forceplates} & \begin{tabular}[c]{@{}l@{}}Force: (N)\\ Moment: (N.m)\\ COP: (m)\end{tabular} & \begin{tabular}[c]{@{}l@{}}Array Format: (150 x 3 x stride)\\ First Dimension: normalized sampled gait over the stride/task (150pt) \\ Second Dimension: x/y/z of variable in global space\\ Third Dimension: stride number (L/R leg have been concatenated)\end{tabular} \\ \hline
\texttt{events} & \begin{tabular}[c]{@{}l@{}}HS: (frame)\\ CutPoints: (frame)\end{tabular} & \begin{tabular}[c]{@{}l@{}}Array Format: (stride x 3)\\ First Dimension: stride number (L/R leg have been concatenated) \\ Second Dimension: start/end/duration of stride in frames (100Hz)\end{tabular} \\ \hline
\end{tabular}
\label{Tab:Normalized}
\end{table}

\end{document}